%% file: iclr2024_conference.tex
\documentclass{article} 
\usepackage{iclr2024_conference,times}

\input{math_commands.tex}

\usepackage{hyperref}
\usepackage{url}

\usepackage{graphicx}
\usepackage{cleveref}
\usepackage{floatrow}
\usepackage{wrapfig}
\usepackage{booktabs, makecell, multirow, tabularx, threeparttable, tabulary}
\usepackage{enumitem}

\title{Revealing data leakage in protein interaction benchmarks}

\author{
    \centerline{
    Anton Bushuiev$^{1\ast}$\qquad
    Roman Bushuiev$^{1,4}\thanks{These authors contributed equally.}$~~\qquad
    Jiri Sedlar$^{1}$\qquad
    Tomas Pluskal$^{4}$
    }\\
    \centerline{\textbf{
    Jiri Damborsky$^{2,3}$\qquad
    Stanislav Mazurenko$^{2,3}$\qquad 
    Josef Sivic$^{1}$
    }}\\
    \centerline{$^{1}$Czech Institute of Informatics, Robotics and Cybernetics, Czech Technical University}\\
    \centerline{$^{2}$Loschmidt Laboratories, Department of Experimental Biology and RECETOX, Masaryk University}\\
    \centerline{$^3$International Clinical Research Center, St. Anne's University Hospital Brno}\\
    \centerline{$^4$Institute of Organic Chemistry and Biochemistry of the Czech Academy of Sciences}
}

\iclrfinalcopy 
\begin{document}

\maketitle

\begin{abstract}
In recent years, there has been remarkable progress in machine learning for protein--protein interactions. However, prior work has predominantly focused on improving learning algorithms, with less attention paid to evaluation strategies and data preparation. Here, we demonstrate that further development of machine learning methods may be hindered by the quality of existing train-test splits. Specifically, we find that commonly used splitting strategies for protein complexes, based on protein sequence or metadata similarity, introduce major data leakage. This may result in overoptimistic evaluation of generalization, as well as unfair benchmarking of the models, biased towards assessing their overfitting capacity rather than practical utility. To overcome the data leakage, we recommend constructing data splits based on 3D structural similarity of protein--protein interfaces and suggest corresponding algorithms. We believe that addressing the data leakage problem is critical for further progress in this research area.
\end{abstract}

\section{Introduction}\label{sec:introduction}

\input{assets/fig_synthetic_splits}

Proteins rarely act alone \citep{berggaard2007methods}, with \textit{protein--protein interactions} (PPIs) playing a critical role in various biological processes, such as cell signaling, metabolism, and gene regulation \citep{alberts2015essential}. These interactions occur when two or more proteins bind together, forming a \textit{protein complex}, and can be either transient or persistent. Over the past decade, there has been a significant advancement in machine learning approaches to better understand protein--protein interactions and to guide wet-lab experiments \citep{geng2019finding}, with tremendous progress in recent years \citep{rogers2023growing}.

The vast majority of protein--protein interactions remain undiscovered. For instance, in the well-studied model organism \textit{Escherichia~coli}, only about half of the estimated interactions have been experimentally identified, and a mere 9\% have known structures \citep{green2021large}. Therefore, biological experts are interested in using machine learning models to make predictions about their PPI case studies of interest \citep{bennett2023improving}. It means that in a practical setting, a machine learning model is expected to generalize to protein interactions most likely not observed in the training data. Therefore, testing machine learning models on protein complexes distinct from training ones is crucial for practically-relevant evaluation.

In this work, we reveal that typical data splits of protein complexes do not satisfy this condition, suffering from significant \textit{data leakage}, i.e., the models are tested on interactions that are nearly identical to those used during training. Specifically, we find that the conventional splitting techniques, which are often based on metadata and sequence similarity, are not sufficient for creating benchmarks to effectively assess generalization beyond training data (\Cref{fig:synthetic_splits}; \Cref{sec:existing-splits}). We further discuss several recent successful examples of data splits (\Cref{sec:splits-structure}) and propose recommendations for splitting PPIs based on \textit{interface} similarity, directly comparing the 3D structures of interacting regions (\Cref{sec:lessons-learned}).

\section{Related work}\label{sec:related-work}

The problem of data leakage has been recently identified and rectified in some other domains such as protein--ligand interactions \citep{buttenschoen2024posebusters, li2023leak} or predicting protein properties \citep{cheng2023accurate, li2020predicting}. Nevertheless, a corresponding analysis for protein interactions is missing. Next, we review the main relevant protein--protein interaction problems.

\paragraph{Protein interface prediction.}

To understand the interactions between two protein structures, machine learning can be employed to predict the pairs of interacting amino acids from each protein. This task, known as protein interface prediction (PIP), has been addressed by various machine learning methods \citep{morehead2023dips, williams2023docknet, du2023split, morehead2022geometric, townshend2019dips, fout2017protein}. In this paper, we argue that further advancements in this area might be hindered by the low quality of sequence-based data splits, which are commonly used as benchmarks.

\paragraph{Protein docking.}

Another challenge is to directly predict the 3D structure of a complex formed by two interacting protein structures. Following initial promising results by \cite{ganea2021independent}, a range of machine learning algorithms have been developed for this problem \citep{yu2024rigid, sverrisson2023diffmasif, du2023split, chu2023flexible, ketata2023diffdock, williams2023docknet}. We find that most research in this area uses leaking benchmarks based on sequence similarity, inherited from the domain of protein interface prediction. To the best of our knowledge, only a single recent study \citep{sverrisson2023diffmasif} rethinks data splits to focus on the similarity of protein interface structures.

\paragraph{Protein binder design.}

Once the structure of a protein complex is known, machine learning can be used to design the binding interface, specifically to predict amino acid mutations that enhance binding strength. This task has been tackled by machine learning for over a decade \citep{geng2019finding}. Initial research used completely random splits of mutations across PPIs, leading to overestimated performance \citep{tsishyn2024quantification}. Here, we show that most recent studies \citep{liu2023predicting, luo2023rotamer, yue2023mpbppi, shan2022deep, liu2021deep} still typically employ benchmarks with high data leakage, based on PDB code splitting. In contrast, \cite{bushuiev2023learning} utilize data splits based on interface similarity. In this workshop paper, we use the iDist method from \cite{bushuiev2023learning} for large-scale PPI comparison, applying it to comprehensively assess data leakage across various tasks.

\paragraph{Other tasks.} Several other tasks related to protein--protein interactions are beyond the scope of this study, particularly those that do not require the structures of protein complexes as inputs. These tasks include, for example, protein complex folding \citep{evans2021protein} and problems related to protein--protein networks \citep{bernett2023cracking, tang2023machine}.

\section{Problems of existing data splits for protein complexes}\label{sec:existing-splits}

In this section, we examine the issue of data leakage prevalent in most of the recently utilized data splits of protein complexes. To quantify the leakage, we calculate the ratio of test PPIs that have structural near duplicates in the training set. For detecting near duplicates, we use the iDist algorithm \citep{bushuiev2023learning}, which enables large-scale structural similarity search for protein--protein interfaces, i.e., directly comparing interacting regions in 3D.

The iDist algorithm efficiently approximates iAlign \citep{gao2010structural} -- the adaptation of traditional TM-align to protein interactions \citep{zhang2005tm}. iDist performs distance-weighted message passing across interface amino acids, aggregating their patterns into real-valued vectors. Each PPI is then represented by a single vector aggregating its structural features over the whole interface. Near-duplicate interfaces are then identified based on  thresholding Euclidean distance between their representing vectors. As shown in~\cite{bushuiev2023learning}, this efficient, alignment-free approximation of the traditional iAlign structural alignment is also accurate. iDist identifies near duplicates found by iAlign with 99\% precision and 97\% recall. 

Next, using iDist we show that splitting PPIs based on metadata (\Cref{sec:splits-metadata}) or sequence similarity (\Cref{sec:splits-sequence}) is insufficient for effectively measuring generalization beyond training interactions. Specifically, we find that these proxy similarity measures result in high ratio of leaked PPIs, as directly measured by the structural similarity of interacting interface regions.

\subsection{Splitting by metadata is not enough}\label{sec:splits-metadata}

\input{assets/fig_skempi_leaks}

The Protein Data Bank (PDB) is a central resource for protein structures \citep{berman2000protein}. With over 200,000 entries, each containing an arbitrary number of interacting proteins, it is common to compare these entries based on their metadata. 

The basic data splitting strategy is to split PPIs randomly, ensuring that the same interactions from the same entries do not appear more than once across data folds \citep{luo2023rotamer, shan2022deep, liu2021deep}. Technically, this is achieved by splitting PPI-specific identifiers (e.g., \texttt{3BTD\_I\_E} for an interaction between protein chains \texttt{E} and \texttt{I} from the \texttt{3BTD} entry in PDB, see \Cref{fig:skempi_leaks}). A more cautious method involves splitting PPIs based on the PDB entry codes from which they originate (e.g., \texttt{3BTD}). This ensures that same interactions from a single symmetric complex (e.g., three nearly identical PPIs from \texttt{1UIJ} in \Cref{fig:canavalin}) remain within the same data fold \citep{morehead2022geometric}. Another alternative is to divide the entries according to their submission date to the PDB, utilizing the most recent entries for testing. Although, to the best of our knowledge, this method has not been applied to PPI data, it is a frequent practice in other domains working with PDB data \citep{corso2022diffdock}, often combined with other similarity criteria \citep{evans2021protein, jumper2021highly}.

Here, we demonstrate that each of the three aforementioned methods alone leads to significant data leakage due to the extensive redundancy within the Protein Data Bank \citep{burra2009global}, as well as the extreme repetitiveness of protein--protein interfaces (\Cref{fig:canavalin}; \cite{gao2010structural, garma2012many}). To assess the data leakage statistically, we first create five random samples of PPIs from PDB, each containing roughly 50,000 interactions. Then, we split each sample according to the aforementioned practices in the $90\%/10\%$ ratio to simulate the train-test splits. Subsequently, we quantify the number of test entries with at least one near-duplicate in the training set, as detected by the iDist algorithm. The results are shown in \Cref{fig:synthetic_splits}.

We find that splits based on PPI codes, on average, lead to $86\%$ data leakage, which is expected due to the high redundancy in PDB. Splits based on PDB codes improve the situation, yet still lead to $65\%$ data leakage. Further refinement of the splits considering the deposition time of the entries leads to a slight improvement, reducing leaks to $61\%$. For more details on the experiment, please refer to \Cref{sec:methods}. In summary, all examined metadata-based splits result in a majority of test PPIs being leaked from the training data, indicating that these naive approaches should not be utilized in practical scenarios.

Finally, we validate our findings by examining data leakage in a recent split \citep{luo2023rotamer} of SKEMPI v2.0 \citep{jankauskaite2019skempi}, a standard benchmark dataset in the field of protein binder design. This approach divides all 343 PPIs derived from the PDB into three sets for three-fold cross-validation and testing based on PPI codes (e.g.,~\texttt{3BTD\_I\_E}; see \Cref{fig:skempi_leaks}). Using iDist, we identify test PPIs that are similar to those in the training or validation sets in all three evaluation scenarios. On average, iDist detects $56\%$ ($57\%$, $57\%$, and $54\%$) of the test PPIs as leaked from the training data (i.e.,~with a near-duplicate interface structure in the training data), despite the small size of the dataset.

\subsection{Splitting by sequence similarity is not enough}\label{sec:splits-sequence}

\input{assets/fig_low_seq_high_interface}

Compared to the field of protein structure, the area of learning from protein sequences is more established, offering advanced tools for sequence comparison and splitting without data leakage \citep{suzek2007uniref}. Specifically, the MMseqs2 algorithm \citep{steinegger2018clustering, steinegger2017mmseqs2} enables ultra-fast large-scale search for similar proteins, and led to the creation of the clustered UniRef database \citep{uniprot2023uniprot}. 

Drawing inspiration from this, the machine learning community has employed sequence similarity methods to split protein complexes, ensuring that the protein sequences in test complexes do not have close homologs in the training data \citep{williams2023docknet, du2023split, dauparas2022robust, morehead2022geometric, townshend2019dips}. Here, we show that even sequence-based splitting of protein complexes can result in a substantial rate of leaked test interactions, i.e., interactions in the test set that have {\em structurally} near-duplicate interactions in the training data, as measured by interface similarity.

Previously, it was demonstrated that the DIPS dataset \citep{townshend2019dips}, when split by protein families to separate similar proteins \citep{ganea2021independent}, suffers from approximately $53\%$ structural data leakage \citep{bushuiev2023learning}. \Cref{fig:low_seq_high_interface} visualizes an example of the leak. 

In this study, we investigate a significantly more strict sequence-based splitting strategy, based on sequence alignment. Following \Cref{sec:splits-metadata}, we assess data leakage using five PPI samples, each containing around 50,000 interactions. First, we employ the \mbox{MMseqs2} algorithm \citep{steinegger2017mmseqs2} to cluster the protein sequences involved in all PPIs so that sequences sharing at least 30\% sequence identity are in the same clusters. The \mbox{MMseqs2} algorithm enables large-scale clustering of protein sequences by reducing sequence alignment only to the pairs that have common fragments, which are identified through a rapid search across the fragment database. After clustering all the sequences, we form groups of PPIs as connected components within a graph, where nodes represent proteins and edges are given either by protein--protein interactions in the dataset or sharing the same \mbox{MMseqs2} cluster. This leads to 5,600 groups of interactions on average, which we subsequently split using the $90\%/10\%$ ratio. 

We find that this sequence-based splitting approach yields a substantial improvement in structural data leakage compared to metadata-based splits, with a leakage rate of $30\%$ (\Cref{fig:synthetic_splits}). Additionally, we perform the same experiment using \mbox{Linclust} \citep{steinegger2018clustering}, the \mbox{MMsesq2} mode enabling even faster clustering, which may be suitable for larger PPI sets available in future. As expected, using \mbox{Linclust} results in a worse average leakage ratio of $37\%$. Overall, although a certain level of test data similarity to the training set is acceptable, sequence-based splitting may still result in nearly identical interfaces and topologies distributed across different folds (\Cref{fig:low_seq_high_interface}).

\section{Best practices for data splitting of protein complexes}\label{sec:splits-structure}

In this section, we review the recent works addressing the data leakage issue, highlighting the best practices toward high-quality splits  and making a step towards evaluation of generalization to new unseen interaction modes, with distinct interfaces.

\subsection{Splitting by interface similarity is recommended}\label{sec:split-interface}

To overcome the limitations of sequence-based protein comparison, \cite{van2023fast} introduced Foldseek---a fast method to compare and cluster structures of single proteins. Foldseek converts consecutive structural patterns of the protein backbone and its contacts into a sequence in the alphabet of structural features and then uses MMseqs2 \citep{steinegger2017mmseqs2} for fast sequence-based search. Some of the most recent works employed Foldseek to compare protein complexes. Specifically, \cite{sverrisson2023diffmasif} and \mbox{\cite{morehead2023dips}} used the method to map residues between two complexes and count the number of matches at the interaction interface regions. While this approach scales well to large PPI data, it may be biased towards comparing whole complexes rather than interacting regions, as Foldseek is designed to match complete protein folds, based on sequential information of the protein backbones. This may be a significant limitation when comparing protein--protein interfaces since protein pairs with different folds may establish similar interfaces \citep{sen2022structural, mirabello2018topology}.

\cite{gainza2023novo} overcame this limitation by directly aligning interacting interface regions using the traditional TM-align method \citep{zhang2005tm}, applied pairwise to all interfaces, which was computationally feasible for their smaller dataset. Finally, \cite{bushuiev2023learning} developed iDist---a scalable approximation of TM-align-based structural alignment methods. This method enabled to construct PPIRef \citep{bushuiev2023learning}, a non-redundant dataset of protein--protein interactions mined from the whole PDB. The iDist algorithm directly compares protein--protein interfaces on a large scale, and can be used to create non-leaking data splits of large PPI sets. Alternatively, when the dataset size does not exceed several thousand interactions, non-leaking splits may also be achieved by employing pairwise comparison of all interfaces using traditional alignment-based tools such as aforementioned TM-align or its adaption to PPIs, iAlign \citep{gao2010ialign}. Overall, directly comparing PPI interfaces addresses the issues of metadata- and sequence-based splitting, and the emerging fast methods enable large-scale preparation of non-leaking splits.

\subsection{Human expertise is highly-beneficial}\label{sec:human-expertise}

It is important to note that some datasets have a natural schema of splitting given by their construction or manual analysis by the authors. For instance, SKEMPI v2.0 \citep{jankauskaite2019skempi}, a standard dataset for PPI mutations, was semi-automatically categorized into distinct groups (such as antibody--antigen or protease--inhibitor interactions) specifically for machine learning purposes. However, this grouping was initially overlooked by the machine learning community in favor of simpler splits based on metadata similarity \citep{bushuiev2023learning, tsishyn2024quantification}. We validate that splitting \mbox{SKEMPI v2.0} based on the domain expertise leads to higher-quality splits. Specifically, when using the train-test splits from \cite{bushuiev2023learning} based on the described domain-expertise grouping, iDist identifies $0\%$ leakage compared to $56\%$ leakage with a PPI code-based split \citep{luo2023rotamer} discussed in \Cref{sec:splits-metadata}.

Another example is BM5 \citep{pang2017deeprank}, a dataset commonly used for training and validating models for scoring PPI docking poses \citep{xu2024deeprank, reau2023deeprank, geng2020iscore}. Although BM5 comprises nearly six million structures, these are synthetic redocked poses of 232 non-redundant complexes, easy to split properly without large-scale analysis. Here, we validate that even the commonly-employed random 10-fold cross-validation split of the poses on the level of originating complexes \citep{xu2024deeprank, reau2023deeprank, renaud2021deeprank} has sufficient quality. Specifically, iDist only detects a single leak ($<1\%$), where one of the interfaces consists of four disjoint parts similar to the other interface. Ignoring the details of the dataset construction in favor of a random split of the six million poses would instead result in near-duplicate redocked structures of same complexes scattered across folds with near-$100\%$ leakage.

\section{Recommendations}\label{sec:lessons-learned}

In this work, we have highlighted the insufficiency of splitting protein complexes based on metadata or protein sequence similarity for effective evaluation, due to the high structural redundancy of the datasets derived from the Protein Data Bank \citep{berman2000protein}. We believe that the abundant issue of data leakage across different addressed tasks may hinder the further advancements in the field. Therefore, we propose the following recommendations for data splitting of protein--protein interactions:

\begin{enumerate}
    \item \textbf{Use interface similarity as the standard criterion for splitting protein interactions}. Directly dividing complexes based on interface similarity addresses the limitations of traditional metadata- and sequence-based methods, yet it has been rarely utilized. Until recently, large-scale comparison of PPIs was infeasible due to the high computational demands of structural alignment. However, recent advancements have led to the development of efficient approximation methods (see \Cref{sec:split-interface}).
    \item \textbf{Thoroughly review the information provided by dataset authors}. Domain experts often provide information beyond the basic inputs and outputs for machine learning models, which can be crucial for developing evaluation strategies that align with practical needs. Neglecting this vital information can trap machine learning research into repeatedly using flawed, leak-prone benchmarks for years (see \Cref{sec:human-expertise}).
    \item \textbf{Quantify and report data leakage when there is no control over train-test splits}. When training from large data or using pre-trained models, the data splitting strategy may become irrelevant or given beforehand. In such scenarios, we recommend employing interface similarity to evaluate and report the extent of overlap between training and test examples. This approach is a standard practice in other machine learning-driven fields, such as computer vision \citep[][Appendix C]{radford2021learning} and natural language processing \citep[][Appendix C]{brown2020language}.
\end{enumerate}

\subsubsection*{Acknowledgments}

This work was supported by the Ministry of Education, Youth and Sports of the Czech Republic through projects e-INFRA CZ [ID:90254], ELIXIR [LM2023055], CETOCOEN Excellence CZ.02.1.01/0.0/0.0/17\_043/0009632, and ESFRI RECETOX RI [LM2023069]. This work was also supported by the European Union (ERC project FRONTIER no.~101097822) and the CETOCOEN EXCELLENCE Teaming project supported from the European Union’s Horizon 2020 research and innovation programme under grant agreement No.~857560. This work was also supported by the Czech Science Foundation (GA CR) through grant 21-11563M and through the European Union’s Horizon 2020 research and innovation programme under Marie Skłodowska-Curie grant agreement No.~891397. Views and opinions expressed are however those of the authors only and do not necessarily reflect those of the European Union or the European Research Council. Neither the European Union nor the granting authority can be held responsible for them.

\bibliography{iclr2024_conference}
\bibliographystyle{iclr2024_conference}

\clearpage

\appendix
\section{Methods}\label{sec:methods}

In this section, we provide technical details of the methods used to quantify similarities between protein--protein interactions. \Cref{sec:methods-comparing} describes the implementation details of the methods, and \Cref{sec:leakage} provides details on randomized train-test splitting experiments (see Figure \ref{fig:synthetic_splits}). The source code\footnote{\url{https://github.com/anton-bushuiev/PPIRef/blob/main/notebooks/revealing_data_leakage_iclr_gem_2024.ipynb}} for the experiments is available as a part of the PPIRef repository for working with 3D structures of protein--protein interactions.

\subsection{Comparing protein--protein interactions}\label{sec:methods-comparing}

\paragraph{Sequence similarity.} 

For illustrative examples, we measure sequence similarity by calculating sequence identity. Specifically, to calculate sequence similarity between two complexes, we take the maximum sequence identity upon the pairwise alignment between all proteins from one complex and all from the other. To align sequences, we follow \cite{buttenschoen2024posebusters} and use \texttt{PairwiseAligner} from Biopython\footnote{\url{https://github.com/biopython/biopython}} \citep{chapman2000biopython} with an open gap penalty of $-11$ and an extension gap penalty of $-1$, as well as the BLOSUM62 substitution matrix. We always normalize sequence identities by dividing by the length of the shorter of the two sequences.

The described pairwise sequence similarity with Biopython, using dynamic programming, is too slow for data splitting experiments with thousands of PPIs. Therefore, to split interactions according to their sequence similarity (see \Cref{fig:synthetic_splits}), we use the MMseqs2 clustering algorithm\footnote{\url{https://github.com/soedinglab/mmseqs2}} \citep{steinegger2017mmseqs2}. Specifically, we run \texttt{mmseqs easy-cluster} with the standard parameters and \texttt{--min-seq-id 0.3} to ensure strict clustering where two proteins with at least $30\%$ sequence identity fall under the same cluster. As discussed in \Cref{sec:splits-sequence}, we also experiment with Linclust \citep{steinegger2018clustering}. We run the algorithm using \texttt{mmseqs easy-linclust} with the default parameters.

\paragraph{Interface similarity.} 

To measure interface similarity across experiments, we use iDist \citep{bushuiev2023learning}, a scalable and accurate approximation of iAlign \citep{gao2010ialign} and US-align \citep{zhang2022us}, two methods based on conventional TM-align \citep{zhang2005tm} for protein structures. We use the iDist implementation from the PPIRef package\footnote{\url{https://github.com/anton-bushuiev/PPIRef}}. We always extract PPI interfaces based on the 6\r{A} contacts between heavy atoms (using the \texttt{PPIExtractor} class). Subsequently, we consider PPIs as near duplicates if their iDist distance (\texttt{IDist.compare}) is lower than 0.04, as reported in \cite{bushuiev2023learning}.

To detect near duplicate PPI interfaces on a large scale for estimating data leakage, we follow the same methodology in a scalable implementation. Specifically, we first embed all interfaces in parallel (\texttt{IDist.embed\_parallel}) and then query the test interfaces against the training interfaces (\texttt{IDist.query}). We consider a test interface leaked if it has at least one near-duplicate hit in the training data.

\subsection{Estimating data leakage}\label{sec:leakage}

To assess the quality of individual methods for train-test splitting, we perform randomized large-scale experiments. In the first step, we extract all 6\r{A} heavy-atom interfaces from PDB as of January 27, 2024, using the aforementioned PPIRef package. We filter out improper interfaces using the standard criteria for buried surface area and structure quality \citep{townshend2019dips, bushuiev2023learning}. This leads to 349,685 interfaces from 99,401 PDB entries.

In the next step, we create 5 subsamples of PPIs by randomly drawing 15,000 PDB codes from the data and selecting all corresponding interactions. This results in 5 sets of roughly 50,000 interactions each. Then, we split each subsample with four different methods discussed in Section \ref{sec:existing-splits} in the approximate 90\%/10\% ratio (20 splits in total), and report the means and standard deviations of their leakage (Figure \ref{fig:synthetic_splits}).

Specifically, for splitting based on PDB (e.g., \texttt{1BUI}) and PPI codes (e.g., \texttt{1BUI\_A\_C}), we simply split the codes in the 90\%/10\% ratio. For splitting based on PDB deposition time, we sort all the PDB codes according to their deposition time and select the most recent 10\% for testing. For splitting based on MMseqs2 clusters, we divide PPIs into groups such that the groups do not share sequences from the same clusters, and split the groups in the 90\%/10\% ratio.

\section{Supplementary figures}\label{sec:figures}

\input{assets/fig_canavalin}

\end{document}

%% file: math_commands.tex
\usepackage{amsmath,amsfonts,bm}

\def\eqref#1{equation~\ref{#1}}

\def\1{\bm{1}}

\DeclareMathAlphabet{\mathsfit}{\encodingdefault}{\sfdefault}{m}{sl}
\SetMathAlphabet{\mathsfit}{bold}{\encodingdefault}{\sfdefault}{bx}{n}



%% file: assets/fig_synthetic_splits.tex
\begin{wrapfigure}{R}{5.5cm}

\vspace{-0.5cm}

\ffigbox[\textwidth]
{
    \caption{\textbf{Data leakage in protein--protein interaction splits.} Bars show the average percentage of test examples having a nearly identical training example for $90\%/10\%$ splits of 50,000 protein--protein interactions from the Protein Data Bank, with standard deviations (error bars) across 5 random samples. Near duplicates are identified using the iDist algorithm.}
}
{
    \label{fig:synthetic_splits}
    \includegraphics[width=0.8\textwidth]{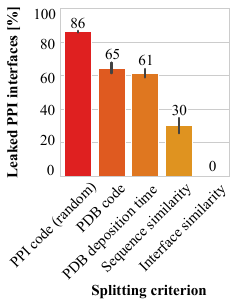}
}

\vspace{-1.15cm}

\end{wrapfigure}

%% file: assets/fig_skempi_leaks.tex
\begin{wrapfigure}{R}{6.5cm}

\vspace{-0.4cm}

\ffigbox[\textwidth]
{
    \caption{\textbf{Splitting by PDB codes causes data leakage in benchmarks for PPI design.} The figure shows three protein complexes taken from SKEMPI v2.0, a standard dataset of annotated PPI mutations. Different chains in the entries are color-coded and labeled with their respective codes. In total, the dataset contains 10 such near-duplicate interactions (PDB codes \texttt{3BTD}, \texttt{3BTE}, \texttt{3BTT}, \texttt{3BTM}, \texttt{3BTQ}, \texttt{3BTW}, \texttt{3BTH}, \texttt{3BTF}, \texttt{3BTG}, \texttt{2FTL}), representing single-point mutants of the same interaction between a serine protease and its inhibitor \citep{krowarsch1999interscaffolding}. Recent machine learning research in protein--protein interactions employed PDB-code splitting, resulting in near-duplicate entries, similar to those shown in this figure, scattered across train-validation-test folds.}
}
{
    \label{fig:skempi_leaks}
    \includegraphics{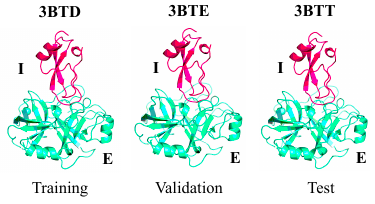}
}

\vspace{-0.5cm}

\end{wrapfigure}

%% file: assets/fig_low_seq_high_interface.tex
\begin{wrapfigure}{R}{6.5cm}

\vspace{-0.3cm}

\ffigbox[\textwidth]
{
    \caption{\textbf{Splitting by sequence similarity introduces data leakage in benchmarks for protein docking and interface prediction.} The figure shows two phosphorylase homooligomers, taken from DIPS, a standard dataset for training and validating machine learning models. The complex to the left (PDB code \texttt{1K3F}), as well as the complex to the right (\texttt{1K9S}), is composed of five identical proteins (highlighted with colors). Nevertheless, the proteins across the entries have very low sequence similarity (26.5\%). Despite the sequences in the complexes being different, the secondary structure of the chains, the topology of the interactions, as well as the 3D structure and the amino acids at the interfaces are highly similar across the entries (iDist $< 0.04$, the near-duplicate threshold; iAlign's p-value $< 10^{-6}$). Recent machine learning research for protein docking and interface prediction employed data splitting based on sequence similarity, resulting in data leakage.}
}
{
    \label{fig:low_seq_high_interface}
    \includegraphics{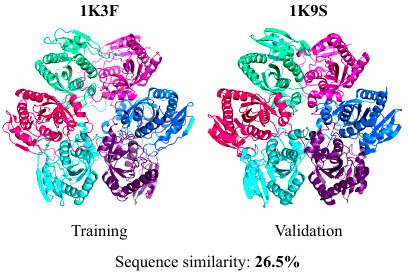}
}

\vspace{-0.75cm}

\end{wrapfigure}

%% file: assets/fig_canavalin.tex
\begin{figure}[h!]
  \centering
  \includegraphics[width=1.0\textwidth]{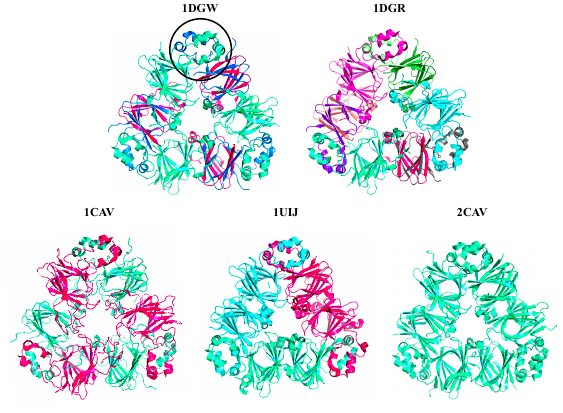}
    \caption[Structural redundancy challenges data splitting of protein interactions.]
    {\textbf{Structural redundancy challenges data splitting of protein interactions.} The figure shows five different PDB entries representing the 3D structure of the canavalin trimer \citep{ko1993three}, with protein chains in different colors. The high level of structural redundancy observed in protein–protein interactions, both within individual entries and across multiple entries, necessitates careful data splitting strategies (for example, the interaction highlighted within the circle is represented in the figure fifteen times, occurring three times in each structure). However, inconsistent metadata and the modular nature of protein structures make naive approaches, such as splitting based on PDB codes or sequence similarity, prone to fail, distributing the same interactions across training-validation-test folds.}
  \label{fig:canavalin}
\end{figure}
